\documentclass[letterpaper, 10 pt, conference]{ieeeconf}  %
\IEEEoverridecommandlockouts                              %

\usepackage{amsmath, amssymb}   %
\usepackage{amsthm}                 %
\newtheoremstyle{italicexample}{}{}{\itshape}{}{\bfseries}{.}{ }{}
\theoremstyle{italicexample}

\theoremstyle{plain}

\usepackage{algorithm}          %
\usepackage{algpseudocodex}      %
\makeatletter
\newcommand\fs@ruledtight{\fs@ruled\def\@fs@post{\kern2pt\hrule\kern-1.5em\relax}}
\makeatother
\floatstyle{ruledtight}
\restylefloat{algorithm}
\usepackage{threeparttable}     %
\usepackage{multirow}           %
\usepackage{booktabs}           %

\usepackage{graphicx}
\usepackage{pifont}
\usepackage{makecell}
\usepackage{xcolor}
\usepackage{annotate-equations}   %
\usepackage{mathtools}
\usepackage{siunitx}
\usepackage[bb=dsserif]{mathalpha}
\usepackage{bm}

\makeatletter
\let\orig@endthebibliography\endthebibliography
\def\endthebibliography{\if@newlist\item[]\fi\orig@endthebibliography}
\makeatother

\makeatletter
\let\orig@makecaption\@makecaption
\long\def\@makecaption#1#2{%
\ifx\@captype\@IEEEtablestring%
\setbox\@tempboxa\hbox{\footnotesize #1:~#2}%
\ifdim \wd\@tempboxa >\hsize%
\parbox[t]{\hsize}{\footnotesize\noindent #1:~#2}%
\else\hbox to\hsize{\footnotesize\hfil\box\@tempboxa\hfil}\fi%
\@IEEEtablecaptionsepspace%
\else\orig@makecaption{#1}{#2}\fi}
\makeatother

\theoremstyle{remark}

\colorlet{satcolor}{orange} %
\colorlet{caucolor}{teal}   %
\usepackage{soul}

\newcommand{\cauterm}[1]{{\sethlcolor{caucolor!\eqnhighlightshade}\hl{#1}}}
\newcommand{\satbox}[2]{\eqnmarkbox[satcolor]{#1}{#2}}
\newcommand{\caubox}[2]{\eqnmarkbox[caucolor]{#1}{#2}}

\newcommand{\part}{$\sim$}

\DeclareUnicodeCharacter{2076}{\textsuperscript{6}}
\DeclareUnicodeCharacter{2077}{\textsuperscript{7}}
\DeclareUnicodeCharacter{1F937}{\raisebox{-0.15em}{\includegraphics[height=1em]{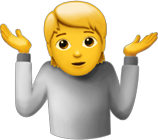}}}  %
\usepackage{xspace}
\newcommand{\method}{CusADi2\xspace}

\newcommand{\para}[1]{\smallskip\noindent\textbf{#1}}
\newcommand{\schedset}{\mathcal{S}_\mathcal{D}} %
\newcommand{\maxwidth}{\operatorname{maxwidth}} %

\colorlet{wincolor}{violet} %
\newcommand{\winbox}[2]{\eqnmarkbox[wincolor]{#1}{#2}}
\newcommand{\winterm}[1]{{\sethlcolor{wincolor!\eqnhighlightshade}\hl{#1}}}
\definecolor{mycitecolor}{RGB}{71, 191, 38}
\definecolor{mylinkcolor}{RGB}{40, 115, 201}
\definecolor{DGreen}{RGB}{107, 190, 35}
\definecolor{darkblue}{rgb}{0.1, 0.1, 1}
\makeatletter
\let\NAT@parse\undefined
\makeatother
\usepackage[nocompress]{cite}  %
\usepackage[colorlinks=true, citecolor=mycitecolor,linkcolor=mylinkcolor,urlcolor=mycitecolor]{hyperref}
\usepackage[noabbrev,capitalise]{cleveref}

\title{\LARGE \bf \method: Latency-Optimized Batch Computation of\\ Robot Kinematics, Jacobians, and Hessians on the GPU}

\author{
John Viljoen$^{1}$,
Lasse Peters$^{1}$,
Masayoshi Tomizuka$^{1}$, and
Negar Mehr$^{1}$
\thanks{$^{1}$J. Viljoen, L. Peters, M. Tomizuka, and N. Mehr are with the Department of Mechanical Engineering, University of California, Berkeley, CA, USA.
        {\tt\small \{john\_viljoen, lasse.peters, tomizuka, negar\}@berkeley.edu}}
\thanks{This work has been submitted to the IEEE for possible publication. Copyright may be transferred without notice, after which this version may no longer be accessible.}
}

\makeatletter
\let\orig@maketitle\@maketitle
\renewcommand{\@maketitle}{\orig@maketitle%
  \par\vspace{-0.5em}%
  {\centering\includegraphics[width=\textwidth]{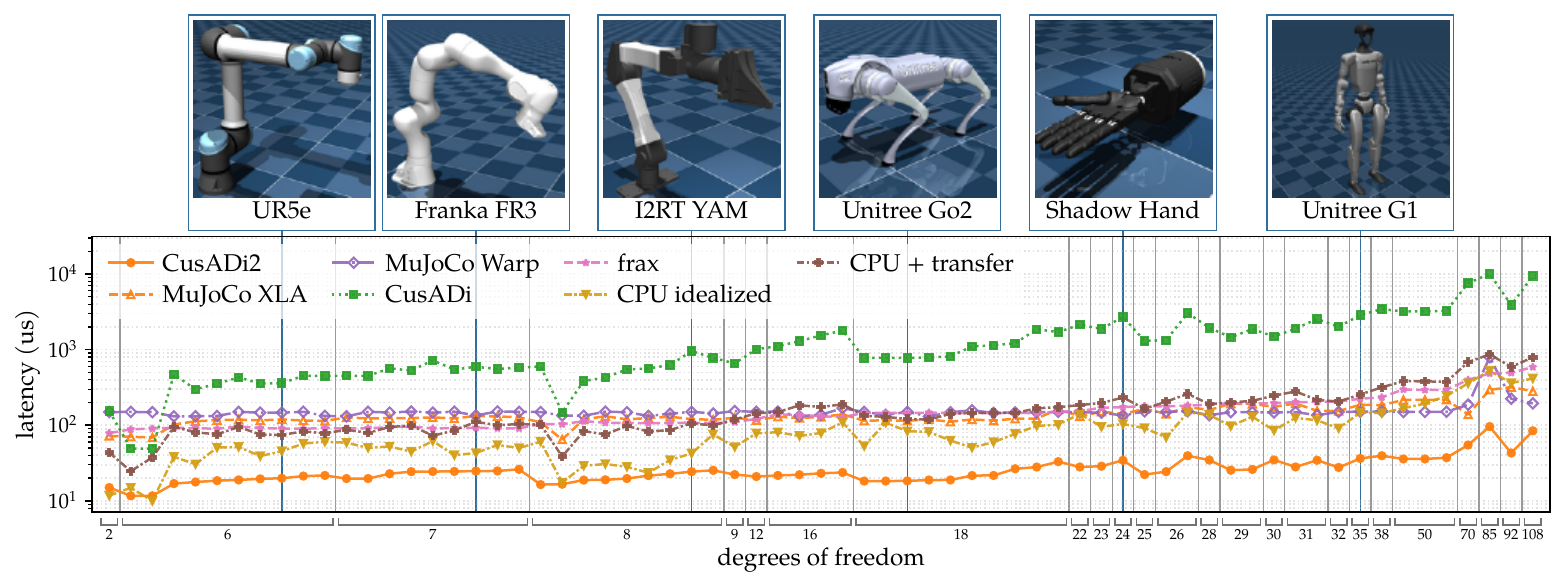}\par}%
  \def\@captype{figure}\refstepcounter{figure}\label{fig:gpu_absolute}%
  \vspace{-10pt}\footnotesize\noindent\@makecaption{\fnum@figure}{
    \method evaluates batches of dynamical-system functions, such as robot kinematics, Jacobians, and Hessians, on the GPU with low latency.
    Here, we show the latency of batched forward kinematics ($B = 256$ rollouts of 8 time steps) for all 67 robots in the MuJoCo Menagerie, sorted by degrees of freedom, with common platforms labeled.
    \method is fastest on all but two low-DOF robots and evaluates common platforms in under 40\,$\mu$s.
    }%
    \vspace{-1.5em}
  }
\makeatother

\begin{document}

\maketitle

\begin{abstract}
Trajectory optimization methods, from sampling-based zero-order to gradient-based first- and second-order approaches, require many evaluations of \emph{dynamical system functions}, i.e., functions derived from a robot's dynamical model, such as forward kinematics and their Jacobians and Hessians.
These evaluations are largely independent, e.g., across sampled trajectories or time steps, and can thus be evaluated in parallel as a \emph{batch}.
When used to make decisions \emph{online}, these methods require \textit{low latency}, i.e., each batch must complete in as little wall-clock time as possible.
Existing frameworks utilize modern GPU hardware primarily to parallelize \emph{across} batch instances, thereby optimizing \emph{throughput} but not \emph{latency}.
We present \method, a software library that reduces the batch evaluation \emph{latency} by additionally parallelizing computation \emph{within} each batch instance.
We benchmark forward kinematics, Jacobians, and Hessians for all 67~robots in the MuJoCo Menagerie, including arms, hands, quadrupeds, humanoids, and biomechanical models, against MuJoCo XLA, MuJoCo Warp, frax, CusADi, and CasADi.
\method achieves lower latency than all baselines at batch sizes of 64 to 4096, while matching them at larger batch sizes.
We also demonstrate the performance advantage of \method in a sampling-based control application.
All code is open-source at: \url{https://github.com/johnviljoen/cusadi2}.%
\end{abstract}

\section{Introduction}

Modern robotics algorithms often make decisions by repeatedly evaluating \emph{dynamical system functions}, i.e., functions derived from a robot's dynamical model, such as forward kinematics and their Jacobians and Hessians \cite{algs_for_decision_making_mykel, massively_parallel_rl}.
In some cases, this dependence on a model is explicit.
For example, model-based online planning techniques, whether zero-order (sampling-based)~\cite{mjpc,mppi} or first- and second-order (gradient-based)~\cite{mpc_textbook}, all repeatedly evaluate a model to predict the effect of their actions on the environment \cite{algs_for_decision_making_mykel}.
In other cases, the dependence on a model may be \emph{implicit}.
For example, model-free techniques such as Reinforcement Learning (RL) usually rely on a simulator to generate sufficient training data \cite{massively_parallel_rl}.
Approaches that evaluate the model \emph{offline}, such as RL, require high throughput. 
Therefore, modern GPU tools \cite{mjx, mjwarp, isaac_lab, cusadi} have been optimized for this criterion.
However, techniques that evaluate the model \emph{online} require a different criterion, namely \emph{low latency}, which remains a secondary concern for these tools.
Sampling-based methods evaluate batches of hundreds of rollouts for every control action, and gradient-based methods evaluate Jacobians and Hessians along the trajectory at every iteration.
In both cases, the latency of these evaluations directly limits the achievable control rate.
These online methods thus directly benefit from low-latency evaluation of dynamical system functions.

In this paper, we fill this gap by introducing \method, a software library for low-latency batched evaluation of dynamical system functions on the GPU.
To be broadly applicable in online planning, \method supports not only forward kinematics but also Jacobians and Hessians.
It exploits the structure of each function's computational graph to parallelize its evaluation.
Two further optimizations reuse one schedule across trajectory stages and group same-type operations, which GPUs execute more efficiently.
It thus achieves lower latency than prior methods such as MuJoCo Warp~\cite{mjwarp}, MuJoCo XLA~\cite{mjx}, and CusADi~\cite{cusadi}.
For 256 forward kinematics evaluations, existing GPU frameworks need at least 100 $\mu$s on nearly every MuJoCo Menagerie~\cite{menagerie} robot, whereas \method needs under 40 $\mu$s on common platforms (\cref{fig:gpu_absolute}).
Across all 67 robots, from dexterous hands to humanoids, this advantage is largest for batches of 64 to 4096, with no throughput loss beyond.
Finally, we demonstrate a \qty{50}{\hertz} safe sampling-based planner for a Franka Panda arm, with a Higher-Order Control Barrier Function~(HOCBF)~\cite{hocbf} safety filter per sampled trajectory.

\section{Related Works}

GPU tools for evaluating dynamical system functions have so far been designed for throughput, as required by RL.
Below, we review these tools along three axes, namely batched simulation, per-instance latency, and derivative sparsity.

\para{Batched GPU Simulation.}
Modern batched GPU simulators~\cite{isaac_lab, brax, mjx, mjwarp, madrona, frax, cusadi} parallelize simulation across thousands of independent instances to generate data at high throughput.
These tools were originally designed for RL, where more environment interactions directly accelerate training~\cite{massively_parallel_rl}.
However, batched simulation is increasingly used beyond RL, often targeting online deployment~\cite{mjpc,mppi_generic}.
Such online applications require \textit{low-latency} batched evaluation, which is of secondary importance to existing GPU simulators but the primary focus of this paper.

\para{Latency.}
Reducing the latency of a batched evaluation requires accelerating each instance, not only parallelizing across instances.
Existing frameworks take one of two approaches to this per-instance parallelism.
Some frameworks~\cite{mjx, brax, frax} rely on the OpenXLA compiler~\cite{xla} from the JAX ecosystem~\cite{jax}, whose heuristics are designed for machine learning workloads dominated by large tensor operations, rather than fine-grained scalar operations that make up dynamical system functions.
Hence, they accelerate each instance only to a small degree.
Other frameworks~\cite{isaac_lab, madrona, mjwarp, mppi_generic} instead require the user to supply a function that is already parallelized per instance, leaving this burden to the practitioner.
In this paper, we introduce a method that automatically parallelizes per-instance computations. %

\para{Sparsity.}
A further limitation of existing batched evaluation frameworks is that they treat the highly sparse Jacobians and Hessians as \textit{dense}. %
\cite{asdex} addresses this with sparse automatic differentiation, but is designed for Jacobians and is inefficient for Hessians.
Other works~\cite{jax2sympy, warpmpc, cusadi} use symbolic differentiation to obtain sparse Jacobians and Hessians, but either their compilation time grows prohibitively with expression size, or they apply no latency reduction techniques~\cite{cusadi}.
Our method inherits the fast sparse symbolic differentiation pipeline of CasADi~\cite{andersson2019casadi} and combines it with latency reduction techniques.

\section{Problem Statement}
\label{sec:problem_statement}

In this paper, we study the problem of evaluating a dynamical-system function (e.g., robot dynamics and their derivatives) at a batch of many input points in parallel on a GPU in minimal time.
This problem arises naturally in sampling-based MPC \cite{mppi, mjpc}, which requires many rollouts over sampled action sequences, and in gradient-based MPC~\cite{mpc_textbook}, where multiple initial guesses can help identify favorable local minima.
To formalize the problem, we first explain why the latency of a batch evaluation hinges on how fast a single batch instance is evaluated.
We then represent the function as a directed acyclic graph (DAG) of atomic operations, abstract the GPU as a set of parallel compute units, and cast minimum-time evaluation of a single instance as the problem of scheduling the DAG's vertices onto these units.
Finally, we return to the batch and state when the resulting schedule is also optimal for the whole batch.

\para{Parallelization Across Batches and Batch Instances.}
A \emph{batch} consists of $B$ evaluations of the same function at different input points, e.g., one per sampled rollout in sampling-based MPC.
We refer to each of these evaluations as a \emph{batch instance}.
Since instances are independent of one another, they can be evaluated concurrently on the GPU. %
A \emph{naive} parallelization strategy would exploit exactly this parallelism \emph{across} instances.
However, the batch computation is complete only once its \emph{slowest} instance finishes, and parallelizing \emph{across} instances does not reduce this per-instance latency.
Instead, we focus on parallelization \emph{within} each instance, which builds on the fact that many of its operations are independent of one another (see the example below).
For the remainder of this section, we therefore ask how to evaluate a single batch instance on a GPU as fast as possible, and return to the per-batch latency once this problem is formalized.

\para{A Minimal Example.}
Before formalizing the problem, we use a minimal example to illustrate how the evaluation of a dynamical-system function can be parallelized.
The key observation is that such functions are not monolithic, but instead decompose into a sequence of ``atomic'' operations, i.e., low-level operations that cannot be further subdivided.
For example, the function
\begin{equation}
    g(x, y) = \sin(x) + \cos(y)
\label{eqn:toy_expr}
\end{equation}
decomposes into the atomic operations $\sin(\cdot)$, $\cos(\cdot)$, and $\cdot+\cdot$.
Given the structure of $g$, some atomic operations are independent of others. The operations $\sin(x)$ and $\cos(y)$ each depend only on the inputs, whereas the sum depends on the results of both.
Given at least two compute units, i.e., two or more processing cores of a simplified GPU, we can therefore compute $g(x, y)$ in only two steps: first, compute $\sin(x)$ and $\cos(y)$ concurrently, then sum them to get the final result.
Our goal is to systematically identify such opportunities for parallelization in general dynamical-system functions, including those arising from complex robots. In fact, in \cref{sec:results}, we benchmark our method on the entire MuJoCo Menagerie~\cite{menagerie}, which includes dynamical-system functions with up to 429,773 operations.

\begin{figure}
    \centering
    \includegraphics[width=1.0\linewidth]{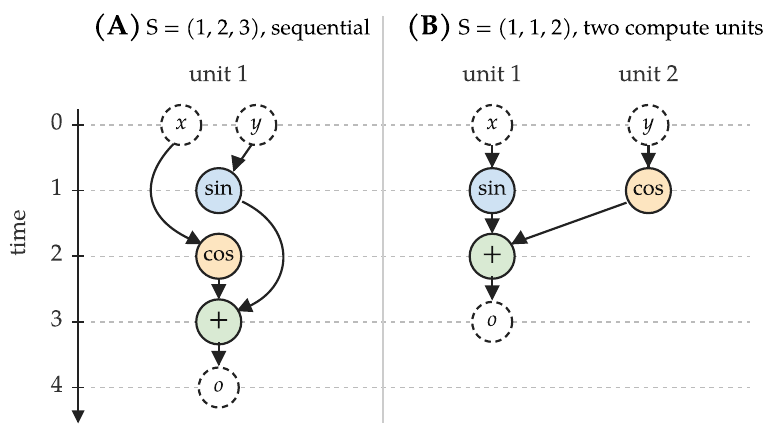}
    \caption{Example schedules for \cref{eqn:toy_expr}.
    \textbf{(A)}: the sequential schedule. \textbf{(B)}: a parallel schedule on two compute units that saves one time step.}
    \vspace{-1.5em}
    \label{fig:scheduler}
\end{figure}

\para{From Function to DAG.}
To formalize minimum-time evaluation of dynamical-system functions, we represent these composite functions as directed acyclic graphs (DAGs):
$$\mathcal{D} = (\mathcal{V}, \mathcal{E}, \mathcal{I}, \mathcal{W}, \mathcal{O}, \operatorname{args}, (f_v)_{v \in \mathcal{W} \cup \mathcal{O}}).$$
In the above DAG $\mathcal{D}$, $\mathcal{V}$ denotes the vertex set, which is partitioned into three pairwise disjoint sets, $\mathcal{V} = \mathcal{I} \cup \mathcal{W} \cup \mathcal{O}$:
\emph{input vertices} $\mathcal{I}$ hold the arguments supplied to the function,
\emph{computational vertices} $\mathcal{W}$ perform atomic operations,
and \emph{output vertices} $\mathcal{O}$ expose the results.
Each non-input vertex $v \in \mathcal{W} \cup \mathcal{O}$ is labeled by a function $f_v: \mathbb{R}^{k_v} \rightarrow \mathbb{R}$.
For computational vertices $w \in \mathcal{W}$, $f_w$ is an atomic operation.
Output vertices $o\in\mathcal{O}$ merely pass through a single value, i.e., $k_o = 1$ and $f_o$ is the identity; thus, only computational vertices incur work.
For instance, the DAG of $g$ in~\eqref{eqn:toy_expr}, shown in~\cref{fig:scheduler}, has input vertices $\mathcal{I} = \{x, y\}$, computational vertices $\mathcal{W} = \{1, 2, 3\}$ with $f_1 = \sin$, $f_2 = \cos$, and $f_3 = +$, and a single output vertex $o$ that passes through the value of vertex~$3$.
The function 
$$\operatorname{args}(v) = (u_1, \ldots, u_{k_v}) \in (\mathcal{I} \cup \mathcal{W})^{k_v}$$
returns the tuple of vertices whose values are input to $f_v$.
For example, for the DAG associated with~\cref{eqn:toy_expr}, we have $\operatorname{args}(3) = (1, 2)$ since the sum (computational vertex 3) takes the results of $\sin(x)$ (computational vertex 1) and $\cos(y)$ (computational vertex 2) as its first and second arguments.
Finally, the edge set $\mathcal{E} \subseteq \mathcal{V} \times \mathcal{V}$ connects each vertex to the vertices it takes as arguments:
$$\mathcal{E} = \{ (u_j, v) \mid v \in \mathcal{W} \cup \mathcal{O},\, u_j \text{ occurs in } \operatorname{args}(v) \},$$
e.g., vertex~$3$ in our example has incoming edges from vertices~$1$ and~$2$.
We require the graph $(\mathcal{V}, \mathcal{E})$ to be acyclic.%

\para{From DAG to Function.}
Next, we show that the DAG fully specifies the function it was built from, so that evaluating the function amounts to evaluating the DAG's vertices in a valid order.
Given inputs $(x_i)_{i \in \mathcal{I}}$, the values of all other vertices are defined recursively by
\begin{align*}
    z_i &= x_i && (i \in \mathcal{I}),\\
    z_v &= f_v(z_{u_1}, \ldots, z_{u_{k_v}}) && (v \in \mathcal{W} \cup \mathcal{O}).
\end{align*}
In our example~DAG for~\cref{eqn:toy_expr}, $z_1 = \sin(x)$, $z_2 = \cos(y)$, and $z_3 = z_1 + z_2$, so the output $z_o = z_3$ recovers $g(x, y)$.
Thus, the DAG represents the function
\begin{equation}
    F_\mathcal{D}:\mathbb{R}^{|\mathcal{I}|} \rightarrow \mathbb{R}^{|\mathcal{O}|}, \quad (x_i)_{i\in \mathcal{I}} \mapsto (z_o)_{o \in \mathcal{O}}.
\end{equation}
For a given function $f$, we denote its associated DAG by $\mathcal{D}_f$, such that $F_{\mathcal{D}_f}\equiv f$.
Henceforth, we use DAG evaluation and dynamical-system function evaluation interchangeably.

\para{Obtaining the DAG.}
In practice, we do not construct~$\mathcal{D}_f$ ourselves but obtain it from CasADi~\cite{andersson2019casadi}, which represents~$f$ as a symbolic expression graph that it evaluates one vertex at a time.
Since CasADi also differentiates symbolically, the Jacobian and Hessian of $f$ are themselves CasADi functions with their own DAGs; hence, our method applies to derivatives (Jacobians and Hessians) without modification.

\para{GPU Abstraction.}
For this problem statement, we abstract the GPU as a collection of $N$ identical compute units.
At each time step, every compute unit independently evaluates one atomic operation associated with a computational vertex, and we assume each operation takes one unit of time.
Although our formulation abstracts away hardware details,\footnote{Examples include the grouping of compute units into warps and thread blocks, the resulting preference for executing the same operation on neighboring units, and data-transfer costs.} our method and implementation explicitly account for these effects; see~\cref{sec:extras}.

\para{Scheduling a DAG Evaluation.}
Our goal is to determine the order in which vertices are evaluated.
To this end, we first observe that the number of time steps any reasonable evaluation requires is bounded by $T := |\mathcal{W}|$, since even the worst-case schedule (evaluating one vertex at a time) finishes within $|\mathcal{W}|$ steps.
With this observation in place, we introduce the notion of a \textit{schedule} $S := (S_1, \ldots, S_{|\mathcal{W}|}) \in \schedset := \{1, \ldots, T\}^{|\mathcal{W}|}$, which assigns each of the $|\mathcal{W}|$ computational vertices a ``time step'' in $\{1, \ldots, T\}$ at which it will be evaluated.
That is, the $w$-th entry, $S_w$, in the tuple $S$ is the time at which the computation of vertex $w \in \mathcal{W}$ is scheduled.\footnote{Input vertices are available from the start and output vertices incur no computation, so neither needs to be scheduled.}
However, we cannot assign these time steps arbitrarily, since a feasible schedule must satisfy two constraints: the \textit{topological order}, which requires each parent to be evaluated before its children, and \textit{hardware saturation}, which limits the number of operations scheduled at any time step to the $N$ available compute units.
Since a schedule depends only on the DAG and on $N$, not on the input values, it is computed once at compile time, i.e., as a one-time offline cost, and then reused for every evaluation.

\para{Width and Depth of a Schedule.}
Two quantities characterize a schedule $S \in \schedset$: \emph{width} and \emph{depth}.
Its \emph{width} at time step $t \in \{1, \ldots, T\}$,
$$\operatorname{width}_t(S) := \sum_{w \in \mathcal{W}} \mathbb{1}[S_w = t],$$
is the number of vertices scheduled at time step $t$, i.e., the number of compute units that $S$ occupies at that step, where $\mathbb{1}[\cdot]$ denotes the indicator function.
Its \emph{depth},
$$\operatorname{depth}(S) := \max_{w \in \mathcal{W}} S_w,$$
is the time step at which the last vertex is evaluated, i.e., the latency of an evaluation that follows $S$.
In these terms, hardware saturation requires $\operatorname{width}_t(S) \le N$ at every time step, and a faster schedule is one of smaller depth.

\para{The Sequential Schedule.}
Tools such as CasADi do not natively identify opportunities for parallel scheduling.
Instead, they evaluate vertices sequentially, one at a time in topological order.
We denote this order by the \emph{sequential schedule} $S^{\text{Seq}} \in \schedset$, which assigns each computational vertex a \emph{unique} time step such that every vertex appears after its parents, i.e., $S^{\mathrm{Seq}}_p < S^{\mathrm{Seq}}_w$ for all $w \in \mathcal{W}$ and $p \in \mathrm{parents}(\mathcal{D}, w) := \{ u \in \mathcal{W} \mid u \text{ occurs in } \operatorname{args}(w) \}$, the computational vertices whose values $w$ consumes.
Since it evaluates one vertex at a time, the sequential schedule has width one at every time step and depth $|\mathcal{W}|$.

\para{Scheduling the Example Function.}
Consider our minimal example from~\cref{eqn:toy_expr} and assume that we have two compute units.
A sequential schedule is shown in~\cref{fig:scheduler}A: with vertices ordered $(\sin, \cos, +)$, $S^{\mathrm{Seq}} = (1, 2, 3)$, i.e., first $\sin(\cdot)$, then $\cos(\cdot)$, and finally $\cdot + \cdot$.
Its width is one at every time step and its depth is three.
A more efficient schedule, $S = (1, 1, 2)$, evaluates $\sin(\cdot)$ and $\cos(\cdot)$ concurrently, then sums their results, as shown in~\cref{fig:scheduler}B.
Its width is two at the first time step and one at the second, so it uses both compute units, and its depth is two, i.e., it saves one time step relative to the sequential schedule.
This raises the question: what is the fastest way to evaluate an arbitrary DAG using at most $N$ compute units?

\para{The Per-Instance Minimum-Time Scheduling Problem.}
Given our DAG and schedule abstractions, we formulate the search for a minimum-time schedule $S_\mathcal{D}^*$, one that evaluates the DAG as quickly as possible using $N$ compute units, as the following optimization problem:
\vspace{1.2em}
\begin{equation}
\begin{aligned}
    S_\mathcal{D}^* \in \operatorname*{arg\,min}_{S \in \schedset} \;\; & \operatorname{depth}(S) \\[0.5em]
    \text{s.t.} \quad & \caubox{cau}{S_p < S_w, \quad \forall\, w \in \mathcal{W},\; p \in \mathrm{parents}(\mathcal{D}, w)} \\[0.5em]
    & \satbox{sat}{\operatorname{width}_t(S) \le N, \quad \forall\, t \in \{1, \dots, T\}}
\end{aligned}
\label{eqn:problem_statement}
\annotate[yshift=1em]{above,right}{cau}{topological order}
\annotate[yshift=-0.8em]{below,left}{sat}{saturation}
\end{equation}
\vspace{0.5em}

\noindent
That is, we seek the schedule of minimum depth among all schedules that satisfy the topological order and saturation requirements discussed above.

\para{Remark: From One Instance to the Batch Schedule.}
A solution $S^*_\mathcal{D}$ of~\eqref{eqn:problem_statement} is a minimum-time schedule for a \emph{single} instance.
Since all instances perform the same operations independently of one another, we can construct a schedule for the entire \emph{batch} by evaluating multiple instances with this same schedule in parallel.
How many instances fit on the GPU at once depends on the maximum width of the instance schedule, $\maxwidth(S^*_\mathcal{D}) := \max_t \operatorname{width}_t(S^*_\mathcal{D})$, i.e., the largest number of compute units the per-instance schedule uses at any time step.
If all instance schedules fit on the GPU at once ($B\,\maxwidth(S^*_\mathcal{D}) \le N$), the entire batch is evaluated in a single \emph{round} and its latency equals that of a single instance, and it is easy to verify that $S^*_\mathcal{D}$ is then also the minimum-time schedule for the batch.
If instead not all instance schedules fit on the GPU at once ($B\,\maxwidth > N$), the batch must be evaluated in multiple rounds, one after another, and the batch latency is that of a single instance times the number of rounds.
When more than one round is needed, per-instance optimality does not necessarily translate to per-batch optimality.\footnote{A narrower but deeper per-instance schedule may fit more instances into one round and thereby finish the batch sooner, and the best batch schedule may even differ across instances. Solving this batch-level problem is harder still and beyond the scope of this paper.}
Despite this theoretical limitation, minimizing per-instance latency leads to significant reductions in batch latency in practice, as we shall see in~\cref{sec:results}.

\section{Method}
\noindent
Finding a per-instance minimum-time schedule for~\eqref{eqn:problem_statement} is NP-hard in general~\cite{np_hard_scheduling_opt}.
However, when the saturation constraint is not binding (i.e., when we have enough compute units to evaluate each vertex as soon as its parents are evaluated), a closed-form solution of~\eqref{eqn:problem_statement} is known~\cite{closed_form_soln_unbounded}.
Fortunately, in many robotics applications the saturation constraint is rarely binding, even on a consumer-grade GPU (RTX~4080).
For example, across all 67 robots of the MuJoCo Menagerie, forward kinematics and Jacobians never saturate, while Hessians saturate for only 4 robots (6.0\%). %
Our method therefore handles two cases: without saturation (\cref{sec:unsaturated}), we solve \eqref{eqn:problem_statement} optimally; with saturation (\cref{sec:saturated}), we use a heuristic that reliably finds a feasible, though potentially suboptimal, schedule.
After discussing each regime, we describe additional method details that leverage the DAG structure and GPU hardware for efficient execution (\cref{sec:extras}).

\subsection{Unsaturated Regime}
\label{sec:unsaturated}

\para{The Level Schedule.}
In the unsaturated regime, we utilize the \textit{level schedule} $S^{\text{L}} \in \schedset$, whose entry for each computational vertex $v \in \mathcal{W}$ is defined recursively as
\begin{equation}
    S^{\text{L}}_v := 1 + \max_{p \in \text{parents}(\mathcal{D}, v)} S^{\text{L}}_p,
    \label{eqn:level_schedule}
\end{equation}
\noindent where the maximum over an empty set is defined to be~$0$, so a vertex whose arguments are all inputs (and which thus has no computational parents) is assigned step~$1$.
Intuitively,~\cref{eqn:level_schedule} schedules every vertex as early as possible, i.e., one step after its latest parent, while respecting the topological order of the DAG.
For our example, $S^{\text{L}} = (1, 1, 2)$, i.e., the efficient schedule in~\cref{fig:scheduler}B is a level schedule for the DAG of~\cref{fig:scheduler}A.
\Cref{alg:level} computes this schedule in a single pass over the vertices of~$\mathcal{D}$ in topological order,\footnote{Obtaining a topological order is trivial: the sequential schedule $S^{\text{Seq}}$ provided by CasADi already is one.} which ensures that every parent is scheduled before its children.

\begin{algorithm}[t]
\caption{Level schedule construction.}
\label{alg:level}
\begin{algorithmic}[1]
\Function{LevelSchedule}{$\mathcal{D}$}
    \State $(w_1, \ldots, w_{|\mathcal{W}|}) \gets \textsc{TopologicalOrder}(\mathcal{D})$ \Comment{parents before children}
    \For{$k = 1, \ldots, |\mathcal{W}|$}
        \State $S^{\text{L}}_{w_k} \gets 1$ \Comment{final if all arguments of $w_k$ are inputs}
        \ForAll{$p \in \text{parents}(\mathcal{D}, w_k)$} \Comment{already scheduled}
            \If{$S^{\text{L}}_p \geq S^{\text{L}}_{w_k}$} \Comment{$w_k$ must run after $p$}
                \State $S^{\text{L}}_{w_k} \gets S^{\text{L}}_p + 1$
            \EndIf
        \EndFor
    \EndFor
    \State \Return $S^{\text{L}}$
\EndFunction
\end{algorithmic}
\end{algorithm}

\para{On the Optimality of the Level Schedule.}
Recall from \cref{sec:problem_statement} that $\operatorname{width}_t(S)$ of a schedule is the number of vertices scheduled at time step $t$ and its depth $\operatorname{depth}(S)$ is the last time step at which a vertex is scheduled.
Intuitively, the wider the schedule, the shallower it becomes, which reduces evaluation latency.
This relationship between width and depth presages the optimality result of~\cite{closed_form_soln_unbounded} for the unsaturated regime.
Specifically, \cite{closed_form_soln_unbounded} proves that, if the width of the level schedule never exceeds the number of compute units, i.e., $\maxwidth(S^{\text{L}}) \leq N$, the level schedule is feasible and optimal for~\eqref{eqn:problem_statement}.

\para{Remark: Depth vs. Computation Time.}
Equating minimal depth with minimal computation time rests on two assumptions: (1) all operations take equal time, and (2) a GPU can execute heterogeneous operations in parallel.
Neither holds exactly. First, operations differ in cost, so the duration of a time step is set by its slowest operation. Second, GPUs support heterogeneous parallel execution only to a limited extent.\footnote{A modern NVIDIA GPU can execute different operations on different warps, but all threads within a warp must execute the same instruction (SIMD).}
We do not model heterogeneous operation costs and leave this to future work, but we address the limitations of GPU heterogeneity in~\cref{sec:extras}.
Moreover, our experiments in~\cref{sec:results} validate the practical effectiveness of our method despite these simplifying assumptions.

\subsection{Saturated Regime}
\label{sec:saturated} 
\noindent
The saturated regime is reached when the maximum width of the level schedule $S^{\text{L}}$ exceeds the number of compute units.
In this regime, finding an optimal schedule is NP-hard~\cite{np_hard_scheduling_opt}.

\para{Windowed Level Schedule.}
To address this, we construct a feasible, though potentially suboptimal, schedule by \emph{windowing} the level schedule.
We cut the sequential schedule into consecutive windows of $K$ vertices each: the first $K$ vertices form window~$1$, the next $K$ form window~$2$, and so on.
Recalling that $S^{\text{Seq}}_w$ is the time step of vertex $w$ in the sequential schedule, window~$b$ thus contains the vertices
\begin{equation}
     \mathcal{W}^K_b := \{\, w \in \mathcal{W} \mid (b-1)K < S^{\text{Seq}}_w \le bK \,\}.
\label{eqn:window}
\end{equation}
We then build a level schedule within each window and evaluate the windows one after another, each starting only once the previous one has finished.
Formally, the windowed level schedule $S^{\text{L},K}$ is given by
\par\noindent
\begin{minipage}{\linewidth}
\vspace{1.5em}
\begin{equation}
\begin{aligned}
    S^{\text{L},K}_w := 1 + \max\Big( & \caubox{caulvl}{\max_{p \in \text{parents}(\mathcal{D}, w)} S^{\text{L},K}_p}, \\[0.3em]
    & \winbox{win}{\max_{u \in \mathcal{W}^K_1 \cup \dots \cup \mathcal{W}^K_{b-1}} S^{\text{L},K}_u} \Big) \quad \text{for } w \in \mathcal{W}^K_b,
\end{aligned}
\label{eqn:windowed_schedule}
\end{equation}
\annotate[yshift=1em]{above,right}{caulvl}{topological order}
\annotate[yshift=-0.8em]{below,right}{win}{window locality}
\vspace{0.5em}
\end{minipage}

\noindent where, as in~\cref{eqn:level_schedule}, we define the maximum over an empty set to be $0$.
Here, \cauterm{the first term enforces the topological order}, as in~\eqref{eqn:level_schedule}; \winterm{the second places $w$ after every vertex of all earlier windows}.
Since each window contains $K$ vertices, choosing $K \le N$ guarantees that $S^{\text{L},K}$ satisfies the saturation constraint of~\eqref{eqn:problem_statement} and is therefore feasible.
\Cref{alg:windowed} computes the windowed level schedule $S^{\text{L},K}$ of~\cref{eqn:windowed_schedule} by running \cref{alg:level} on each window in turn.
Specifically, it applies \cref{alg:level} to the sub-DAG induced by window $\mathcal{W}^K_b$, in which vertices of earlier windows become inputs, denoted by $\mathcal{D}[\mathcal{W}^K_b]$.
This is valid because all earlier windows have finished when a window starts; the window's level schedule is then shifted by the deepest time step $d$ assigned so far.

\begin{algorithm}[t]
\caption{Windowed level schedule construction.}
\label{alg:windowed}
\begin{algorithmic}[1]
\Function{WindowedLevelSchedule}{$\mathcal{D}$, $K$}
    \State $(w_1, \ldots, w_{|\mathcal{W}|}) \gets \textsc{TopologicalOrder}(\mathcal{D})$ \Comment{order of $S^{\text{Seq}}$}
    \State $d \gets 0$ \Comment{deepest time step assigned so far}
    \For{$b = 1, \ldots, \lceil |\mathcal{W}| / K \rceil$}
        \State $\mathcal{W}^K_b \gets (w_k)_{k = (b-1)K + 1}^{\min(bK, |\mathcal{W}|)}$ \Comment{window $b$}
        \State $S^{\text{L}} \gets \textsc{LevelSchedule}(\mathcal{D}[\mathcal{W}^K_b])$ \Comment{earlier windows count as inputs}
        \ForAll{$w \in \mathcal{W}^K_b$}
            \State $S^{\text{L},K}_w \gets d + S^{\text{L}}_w$ \Comment{start after all earlier windows}
        \EndFor
        \State $d \gets \max_{w \in \mathcal{W}^K_b} S^{\text{L},K}_w$
    \EndFor
    \State \Return $S^{\text{L},K}$
\EndFunction
\end{algorithmic}
\end{algorithm}

\para{Choosing the Window Size.}
The window size interpolates between two extreme scheduling strategies.
With $K \ge |\mathcal{W}|$, we obtain a single window, and it is easy to verify that~\cref{eqn:windowed_schedule} reduces to the level schedule, i.e., $S^{\text{L},K} = S^{\text{L}}$.
With $K = 1$, every window is a single vertex and \cref{eqn:windowed_schedule} reduces to the sequential schedule, i.e., $S^{\text{L},K} = S^{\text{Seq}}$.
Since scheduling happens once at compile time, we can afford to search for a good $K$ offline.
In practice, we identify a suitable window size~$K$ by running \cref{alg:windowed} for candidate window sizes $K \in \{|\mathcal{W}|, |\mathcal{W}|/2, |\mathcal{W}|/4, \ldots\}$ and keeping the first window size that produces a schedule whose maximum width does not exceed the number of available compute units~$N$. %
In the \emph{unsaturated regime}, this search returns $K \ge |\mathcal{W}|$ and the windowed level schedule reduces to the optimal level schedule.
In the \emph{saturated regime}, it yields a feasible, though potentially suboptimal, schedule.

\subsection{Additional Considerations for Efficient Execution}
\label{sec:extras}
\noindent
Although the scheduling strategy outlined above is already effective, we make two further improvements that take into account the DAG structure and the GPU hardware.

\para{Staging.}
Many functions in online decision making, such as MPC, are \textit{stage-structured}: they apply the same function~$h$ (e.g., the dynamics and their derivatives) independently at each of the $H$ stages of a trajectory.
The DAG of such a function therefore consists of $H$ disconnected copies of the DAG of~$h$, i.e., it is simply a wider DAG.
\method detects this repeated structure and schedules the DAG of the per-stage function~$h$ only once; all $H$ stages then reuse this schedule, so the scheduling work does not grow with the number of stages.

\para{Vertex Ordering.}
Our scheduling approach models a GPU as a collection of compute units, each of which can evaluate any atomic operation independently.
On modern NVIDIA GPUs, however, compute units are grouped into so-called \emph{warps} of 32~units that must all execute the same operation simultaneously (i.e., in ``lockstep''). %
Consequently, if the units of a warp require $k$ distinct operations, the warp executes these operations sequentially, and each unit discards the results of the operations it does not need.
Thus, a warp with~$k$ distinct operations requires~$k$ passes instead of one, an effect known as \textit{warp divergence}.
To mitigate warp divergence, we sort the vertices within each level by operation type.
For example, consider a level with 32 $\sin(\cdot)$ and 32 $\cos(\cdot)$ vertices that are assigned to two warps.
If the operations alternate ($\sin, \cos, \sin, \cos, \ldots$), each warp contains both operations and thus requires two passes.
Sorting the vertices by operation instead assigns all $\sin(\cdot)$ vertices to the first warp and all $\cos(\cdot)$ vertices to the second, so that each warp completes in a single pass and the evaluation time of this level is halved.\footnote{In principle, warp divergence could be eliminated entirely via \textit{warp specialization}, but at the cost of lower GPU utilization and a more complex schedule assembly.}
This reordering is valid because, by construction, the vertices within a level are mutually independent and can therefore be evaluated in \emph{any} order.

\section{Function-Level Benchmarks}
\label{sec:results}
\label{sec:results_functions}
\noindent
Online decision-making algorithms such as sampling-based and gradient-based MPC spend a large amount of their computation time evaluating the dynamical system and its derivatives, so the latency of these evaluations bounds how fast the overall algorithm can run.
We therefore first study these evaluations in isolation, decoupled from any particular downstream algorithm.
This lets us attribute latency differences directly to the evaluation backend, and lets us test on a much broader set of robots than any single application would allow.
We design these experiments to answer two questions:
\begin{enumerate}
    \item[\textbf{Q1}] How does the latency of \method compare to existing GPU and CPU frameworks across batch sizes? (\cref{sec:results_latency})
    \item[\textbf{Q2}] Which dynamical system functions benefit most from level parallelism, and why? (\cref{sec:results_shape})
\end{enumerate}

\para{Benchmark Setup.}
We evaluate forward kinematics, Jacobians, and Hessians for all 67 robots in the MuJoCo Menagerie \cite{menagerie}, which yields 201 functions with up to 429,773 operations.
Each batch instance evaluates the function along a rollout of 8 time steps, and we test batch sizes $B$ from $1$ to $16384$ at fp32 precision. %

\para{Baselines.}
We compare against four GPU frameworks: MuJoCo XLA \cite{mjx}, MuJoCo Warp \cite{mjwarp}, CusADi \cite{cusadi}, and frax \cite{frax}. 
The CPU framework we test against is CasADi's C code generation \cite{andersson2019casadi}.
We report the CPU baseline both as compute alone (CPU idealized) and with GPU $\rightarrow$ CPU $\rightarrow$ GPU data transfer overhead that a practioner would face if this were integrated into a GPU pipeline (CPU + transfer).
All tests are run on a single machine with an AMD Ryzen 5800X3D CPU with \qty{32}{\giga\byte} RAM and an RTX 4080 GPU with \qty{16}{\giga\byte} VRAM.
All CPU tests run 16 parallel instances on all 16 threads of our CPU. All tests are repeated 5 times, and the mean is used in all analyses.

\subsection{How Does \method Latency Compare to Baselines?}
\label{sec:results_latency}
\begin{figure*}[t]
    \centering
    \includegraphics[width=\textwidth]{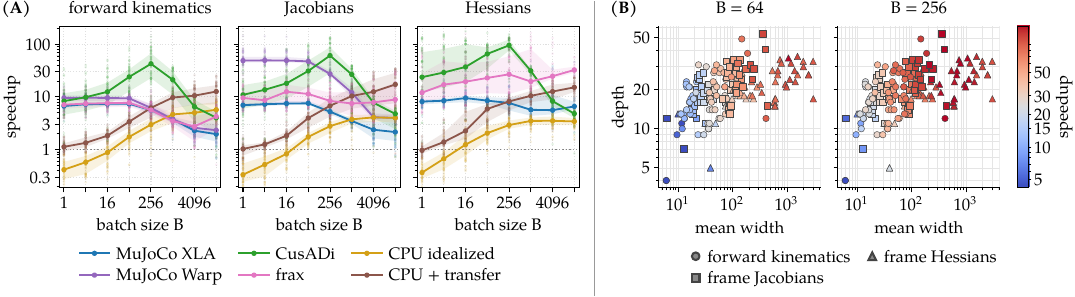}
    \vspace{-2em}
    \caption{We compare the latency of \method against each baseline across the MuJoCo Menagerie. \textbf{(A)}: we plot the speedup of \method over each baseline as a function of batch size. Speedup is the ratio of the two backends' wall-clock latencies, each averaged over 5 runs. Each faded dot is one robot, the solid line is the median across robots, and the shaded band spans the 25th to 75th percentile. \textbf{(B)}: we show the speedup of \method over CusADi for every robot at two batch sizes. Each robot is placed by the mean width and maximum depth of its DAG and colored by its speedup.}
    \vspace{-1.5em}
    \label{fig:speedup_vs_batch}
\end{figure*}

\para{Protocol.}
To answer \textbf{Q1}, we test the forward kinematics, Jacobians, and Hessians of \textit{every} robot in the MuJoCo Menagerie across all baselines.
We evaluate these functions across batch sizes $B \in \{2^{2k} \mid k = 0, 1, \ldots, 7\}$ and measure the wall-clock time from reading the inputs to writing the outputs. %
We plot this result in \cref{fig:speedup_vs_batch}A. 

\para{Largest Batch Results.}
At the largest batch size ($B = 16384$), the batch dimension alone provides enough parallelism to saturate the GPU, so all GPU frameworks outperform the CPU baselines on forward kinematics.
Consequently, the additional parallelism exposed by the level schedule matters less, and the speedup of \method over the other GPU frameworks is small, e.g., to roughly $2\times$ over MuJoCo XLA and MuJoCo Warp for forward kinematics.
Nevertheless, \method remains faster than every baseline at this batch size and still shows a distinct advantage for Jacobians and Hessians.

\para{Intermediate Batch Results.}
The advantage of \method is most pronounced in the intermediate batch regime ($64 \leq B \leq 4096$), which covers the batch sizes common in online decision making, such as the number of sampled trajectories in sampling-based MPC \cite{sampling_mpc}.
Here, the median speedup over CusADi peaks at $B = 256$ at roughly $40\times$, $60\times$, and $95\times$ for forward kinematics, Jacobians, and Hessians, respectively, and \method remains several times faster than MuJoCo XLA, MuJoCo Warp, and frax throughout this regime.
The reason is that frameworks which parallelize only across the batch leave most of the GPU idle until the batch is large, whereas the level schedule additionally exposes the parallelism within each function and thus saturates the GPU at much smaller batch sizes.
Consequently, \method overtakes even the idealized CPU baseline at $B \geq 64$, whereas for forward kinematics the other GPU frameworks do so only around $B = 1024$ (MuJoCo XLA, MuJoCo Warp, frax) or beyond $B = 4096$ (CusADi), and for Jacobians and Hessians several of them never beat the CPU baselines at all within the tested range.
In practice, this means an online algorithm benefits from utilizing \method for dynamical system function evaluations starting at around 64~samples.

\para{Small Batch Results.}
At small batch sizes ($B < 16$), the idealized CPU baseline outperforms \method.
Individual GPU cores are considerably slower than CPU cores, so GPUs excel only through massive parallelism.
Here, however, the CPU already runs at most one instance per thread, while most GPU cores sit idle.
Once transfer is included, \method matches or outperforms the CPU baseline at $B \in \{1, 4, 16\}$ on average.
This is the relevant comparison whenever the surrounding pipeline already runs on the GPU (e.g., sampling, cost evaluation, or a learned policy), since falling back to the CPU at small batch sizes incurs exactly this transfer.
The idealized CPU baseline is therefore preferable only for purely CPU-based pipelines at small batch sizes, where existing code generation such as CasADi already suffices.\footnote{A GPU-only workflow has the additional benefit of leaving the CPU free for other tasks during rollouts.}

\para{Key Takeaways.}
\method achieves the lowest latency of all frameworks in the intermediate batch regime ($64 \leq B \leq 4096$) while remaining competitive at both extremes: at small batch sizes, it matches the CPU baseline once GPU transfer is accounted for, and at large batch sizes, it remains faster than existing GPU frameworks.

\subsection{Which Robots Benefit the Most?}
\label{sec:results_shape}
\noindent
Although \cref{fig:speedup_vs_batch}A captures the general trends, it does not tell a practitioner whether \method suits the particular robot they wish to simulate~(\textbf{Q2}).

\para{Per-Robot Latency Across Degrees of Freedom.}
\Cref{fig:gpu_absolute} reports the absolute latency across the MuJoCo Menagerie at $B = 256$, a regime common in online decision making~\cite{sampling_mpc}.
\method achieves the lowest latency for all robots except two with very few degrees of freedom, where the idealized CPU baseline is marginally faster.
For common platforms such as the UR5e, Franka FR3, Unitree Go2, and Unitree G1, \method evaluates a full batch in under 40\,$\mu$s, roughly 2--5$\times$ faster than the fastest baseline.
Moreover, the latency of \method grows only mildly with the degrees of freedom, whereas that of CusADi, which follows the sequential schedule and only parallelizes across the batch dimension, grows by more than an order of magnitude.
However, degrees of freedom alone do not determine latency.
For example, \method evaluates the 92-DOF IIT SoftFoot more than twice as fast as the 85-DOF MS Human 700 model.
This counterintuitive relationship between degrees of freedom and latency motivates our next analysis.

\para{Speedup by DAG Shape.}
To understand which robots benefit most from \method, \cref{fig:speedup_vs_batch}B reports its speedup over CusADi at two batch sizes, $B \in \{64, 256\}$.
Each marker corresponds to the DAG of one dynamics function (forward kinematics, Jacobian, or Hessian) of one Menagerie robot, placed according to its depth and mean width and colored by speedup.
Since CusADi evaluates the sequential schedule directly, this comparison isolates the effect of our level schedule.
The speedup grows primarily with mean width, which measures the parallelism available to the level schedule.
Depth, in contrast, has little effect on the speedup and instead determines the absolute latency.\footnote{For example, a snake-like robot (a single long kinematic chain) yields a deep, narrow DAG with little opportunity for parallelism, whereas a spider-like robot (eight short, independent legs) yields a shallow, wide DAG with ample parallelism.}

\para{Key Takeaways.}
The benefit of \method grows with the parallelism a function exposes at each level, not with its size: robots whose kinematic trees branch into many independent chains, such as multi-legged robots, humanoids, and dexterous hands, gain the most, whereas long serial chains gain the least.

\section{Downstream Application}
\label{sec:results_mppi}
\noindent Having characterized latency in isolation, we now ask:
\begin{enumerate}
    \item[\textbf{Q3}] How well do \method's gains transfer to a downstream robotics task?
\end{enumerate}
To answer Q3, we adopt a task from \cite{frax} where a Franka arm reaches successive end-effector waypoints while respecting joint limits and avoiding singularities, self-collisions, the floor, and a spherical obstacle between waypoints. %

\para{Controller.}
We solve this task with the predictive-sampling planner of MuJoCo MPC \cite{mjpc}, a sampling-based MPC method, applied to the Franka dynamics with an HOCBF safety filter \cite{hocbf} on its input.
The planner thus optimizes directly over the safety-filtered dynamics and only proposes trajectories that are safe under the model.
Safety-filtering the candidate rollouts of sampling-based controllers is an active area of research \cite{gsmppi, exactmppi}, which makes it a relevant application for \method.
Each control step runs three planner iterations with $B$ samples, each a cubic spline of joint targets with seven control points over a \qty{0.7}{\second} horizon.

\para{Baselines.}
We instantiate this controller with each method from~\cref{sec:results} for evaluating the safety-filtered dynamics, i.e., the HOCBF filter together with the Franka dynamics.
Since the HOCBF constraints require second-order derivatives of the forward kinematics, we exclude MuJoCo Warp, which does not support the computation of Hessians.

\begin{figure}[t]
    \centering
    \includegraphics[width=1.0\linewidth]{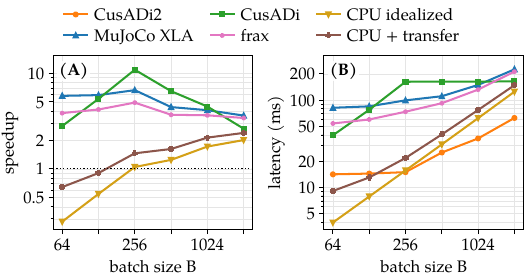}
    \vspace{-2em}
    \caption{
    Comparison of \method against all baselines on the safe predictive sampling task.
    \textbf{(A)}: Speedup of \method over each baseline as a function of batch size on the safe predictive sampling task. \textbf{(B)}: Absolute latency of one control step for all backends.}
    \vspace{-1.5em}
    \label{fig:mppi_cbf_speedup}
\end{figure}

\para{Results.}
\Cref{fig:mppi_cbf_speedup} reports latency, both relative and absolute, averaged across five runs of the safe predictive sampling task.
The end-to-end results show similar performance trends to those of \cref{sec:results_functions}:
\method outperforms all GPU baselines at every batch size tested and is outperformed by both CPU baselines only at smaller batch sizes.

\section{Conclusion}
\label{sec:conclusion}

We presented \method, a software library that reduces the latency of batched dynamical system function evaluations on the GPU by parallelizing computation not only across but also within batch instances.
Across the forward kinematics, Jacobians, and Hessians of all 67 robots in the MuJoCo Menagerie, \method is faster than the existing GPU frameworks we tested at every batch size, and it overtakes an idealized CPU baseline from about 64 batch instances onward.
The gains are largest for robots whose kinematic trees branch into many independent chains, and they carry over to a safe sampling-based MPC controller.
Our scheduling model currently assumes that all operations take equal time.
Future work could reduce latency further by accounting for per-operation costs and by making the schedule more hardware-aware, e.g., via warp specialization~\cite{warp_specialization}.

\addtolength{\textheight}{-3cm}   %

\bibliographystyle{IEEEtran}
\bibliography{lib}

\end{document}